\documentclass{article}

\usepackage{times}
\usepackage{graphicx} 
\usepackage{subfigure} 

\usepackage{natbib}

\usepackage{algorithm}
\usepackage{algorithmic}

\usepackage{hyperref}

\usepackage[accepted]{icml2017}

\icmltitlerunning{Attentive Recurrent Comparators}

\begin{document} 

\twocolumn[
\icmltitle{Attentive Recurrent Comparators}



\icmlsetsymbol{equal}{*}

\begin{icmlauthorlist}
\icmlauthor{Pranav Shyam}{rvce,iisc}
\icmlauthor{Shubham Gupta}{iisc}
\icmlauthor{Ambedkar Dukkipati}{iisc}
\end{icmlauthorlist}

\icmlaffiliation{rvce}{Department of Computer Science and Engineering, Rashtreeya Vidyalaya College of Engineering, Bengaluru, India}
\icmlaffiliation{iisc}{Department of Computer Science and Automation, Indian Institute of Science, Bengaluru, India}

\icmlcorrespondingauthor{Pranav Shyam}{pranavshyam13@gmail.com}

\icmlkeywords{Deep Learning, ICML, attention, recurrent neural networks, one shot learning}

\vskip 0.3in
]



\printAffiliationsAndNotice{} 

\begin{abstract} 
Rapid learning requires flexible representations to quickly adopt to new evidence. We develop a novel class of models called Attentive Recurrent Comparators (ARCs) that form representations of objects by cycling through them and making observations. Using the representations extracted by ARCs, we develop a way of approximating a \textit{dynamic representation space} and use it for one-shot learning. In the task of one-shot classification on the Omniglot dataset, we achieve the state of the art performance with an error rate of 1.5\%. This represents the first super-human result achieved for this task with a generic model that uses only pixel information.

\end{abstract} 

\section{Introduction}
\label{intro}

Utilizing the success and the potential of Deep Neural Networks to solve hard Artificial Intelligence tasks requires neural models that are capable of performing rapid learning ~\citep{DBLP:journals/corr/LakeUTG16}. For models to embody such rich learning capabilities, we believe that a crucial characteristic will be the employment of \textit{dynamic representations} -- representations that are formed by observing a growing and continually evolving set of  features. We call the space that is formed by such evolving representations the \textit{dynamic representation space}. 

In this paper, we present a novel model for one-shot learning that utilizes a crude approximation of such a dynamic representation space. This is done by constructing the representation space lazily and relative to a particular (test) sample every time. For the purpose of producing such relative representations, we develop a novel class of models called Attentive Recurrent Comparators (ARCs).

We first test ARCs across many tasks that require assessment of visual similarity. We find that ARCs that do not use any convolutions show comparable performance to Deep Convolutional Neural Networks on challenging datasets like CASIA WebFace and Omniglot. Though dense ARCs are as capable as ConvNets, a combination of both ARCs and convolutions (ConvARCs) produces much more superior models. In the task of estimating the similarity of two characters from the Omniglot dataset, ARCs and Deep ConvNets both achieve about 93.4\% accuracy, whereas ConvARCs achieve 96.10\% accuracy. In the task of face verification on the CASIA Webface dataset, ConvARCs achieved 81.73\% accuracy surpassing the 79.48\% accuracy achieved by a CNN baseline considered. 

We then use ARCs as a means for developing a lazy, relative representation space and use it for one-shot learning.  On the challenging Omniglot one-shot classification task, our model achieved an accuracy of 98.5\%, significantly surpassing the current state-of-the-art set by all other methods. This is also the first super-human result achieved for this task with a generic model that uses only pixel information.

\subsection{Comparing Objects}
ARCs are inspired by our interpretation of how humans generally compare a set of objects. When a person is asked to compare two objects and estimate their similarity, the person does so by repeatedly looking back and forth between the two objects. With each glimpse of the object, a specific observation is made. These observations which are made in both objects are then cumulatively used to come to a conclusion about their similarity. A crucial characteristic of this process is that new observations are made conditioned on the previous context that has been investigated so far by the observer. The observation and it's contextual location are all based on intermediate deductions -- deductions that are themselves based on the observations made so far in the two objects. A series of such guided observations and their entailing inferences are accumulated to form a final the judgement on their similarity. We will refer to how humans compare objects as the \textit{human way}.

In stark contrast to this, current similarity estimating systems in Deep Learning are analogues of the Siamese similarity learning system~\citep{bromley1993signature}. In this system, a fixed set of features is detected in both the objects. The two objects are compared based on mutual agreement of the detected features. More concretely, comparison between two objects in this system consists of measuring the distance between their vector embeddings or representations. A neural network that is specifically trained to detect the most salient features in an object for a task defines the object to embedding mapping. Detection of features in one object is independent of the features present in the other object.

There is a major underlying difference between the human approach discussed above and the siamese approach to the problem. In the \textit{human way}, the information from the two objects is fused from the very beginning and this combined information primes the subsequent steps in comparison. There are multiple lookups on each of the objects and each of these lookups are conditioned on the observations of both the objects so far. In the \textit{siamese way}, when the embeddings are compared the information fuses mostly at an abstract level and only in the last stage. 

Inspired by the human way, we develop an end-to-end differentiable model that can learn to compare objects called Attentive Recurrent Comparators (ARCs).

Fundamentally, the excellent performance of ARCs shows the value of "early fusion" of information across the context and the value of dynamic representations. Further, it also gives merit to the view that attention and recurrence together can be as good as convolutions in a few special cases. 

Finally, the superior similarity learning capability of ARCs can be of independent interest as an alternative to siamese neural networks for tasks such as face recognition and voice verification.

\begin{figure}[ht]
\vskip 0.2in
\begin{center}
\centerline{\includegraphics[width=\columnwidth]{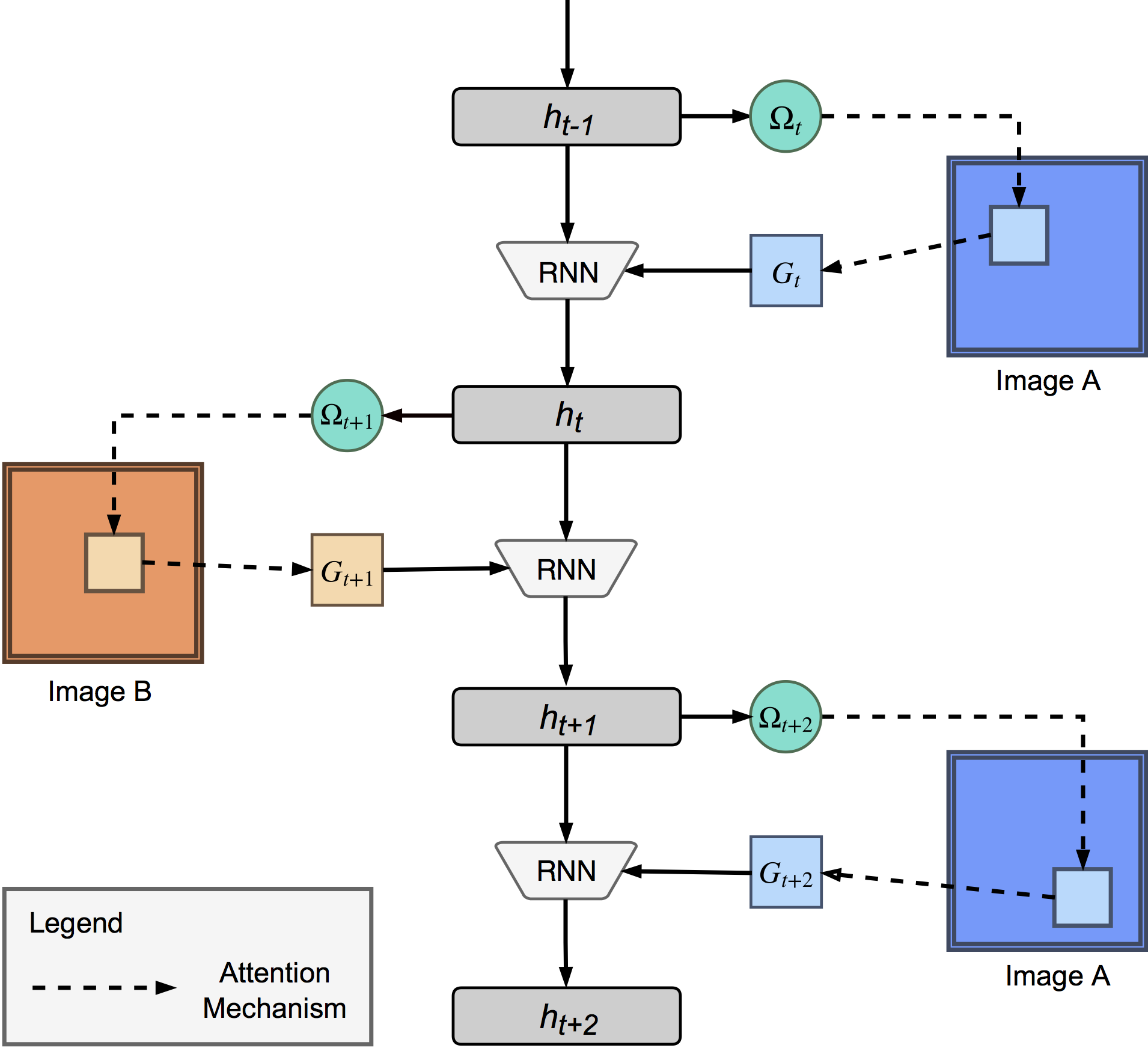}}
\caption{The abstract graph of an ARC comparing two images. The controller which is an RNN primes the whole process. The two images are alternatively and repeatedly attended to. At each time-step the glimpse taken from the image is based on the attention parameters $\Omega_t$ which is calculated using the previous state of RNN $h_{t-1}$ by projecting it with $W_g$. The glimpse obtained $G_t$ and the previous state $h_{t-1}$ together used to update the state of controller to $h_t$.}
\label{ARC Model}
\end{center}
\vskip -0.2in
\end{figure}  

\section{Attentive Recurrent Comparators}
\label{ARC_model}

Our ARC model is essentially an algorithmic imitation of the human way discussed in Section 1.1 and built with Deep Neural Networks. Using attention and recurrence, an ARC makes an observation in one object conditioned on the observations made so far in both objects. The exposition of an ARC model that can compare two images and judge their similarity is given below. But it can be trivially generalised to more images or other modalities.

The model consists of a recurrent neural network controller and an attention mechanism that takes in a specially constructed presentation sequence as the input. Given two images $\{x_a, x_b\}$, we alternate between the two images for a finite number of presentations of each image to form the presentation sequence $x_a, x_b, x_a, x_b, ... , x_a, x_b$. The model repeatedly cycles through both the images, attending to one image at one time-step. 

For time-step $t$ the image presented is given by:
\begin{center}
$I_t  \longleftarrow x_a \:$if $t$  \% $2$ is $0\;$ else $x_b$
\end{center}

The attention mechanism focuses on a specific region of the image current image $I_t$ to get the glimpse $G_t$.
\begin{center}
$G_t \longleftarrow attend(I_t, \Omega_t) \qquad$ where $\quad \Omega_t = W_gh_{t-1}$
\end{center}
$attend(.)$ is the attention mechanism that acts on image $I_t$ (described in the Section 2.1). $\Omega_{t}$ are the attention glimpse parameters which specify the location and size of the attention window. At each step, we use the previous hidden state of the RNN controller $h_{t-1}$ to compute $\Omega_{t}$. $W_g$ is the projection matrix that maps the hidden state to the required number of attention parameters.

Next, both the glimpse and previous hidden state are utilized to form the next hidden state. 
\begin{center}
$h_t \longleftarrow RNN(G_t, h_{t-1})$
\end{center}
$RNN(.)$ is the update function for the recurrent controller being used. This state update function could either be simple RNN or an LSTM. 

Over the course of many time steps, model observes many aspects of both the images. The observations are made by the model at each time step by directing its attention to a region of interest in each input. Since the controller of the model is a Recurrent Neural Network, this round robin like cyclic presentation of images allows for early fusion of information from both images. This makes the model aware of the context in which it is operating under. Consequently, this provides feedback to the attention mechanism to attend on the relevant and crucial parts of each image considering the observations made so far in both the images.

If we make $g$ glimpses (or observations) of each image, the hidden state of the RNN controller at the final time-step $h_T = h_{2g}$ can then be used as the relative representation of $x_a$ with respect to $x_b$ or vice versa. Note that $I_t$ for some $t$ alternates between $x_a$ and $x_b$, while the rest of the equations are exactly the same for all time steps. 

We arrived at the iterative attention paradigm after trying out many approaches to attend to multiple images at once on a few toy datasets. Other approaches for early fusion like attending to both images in the same time-step or having 2 controllers with shared weights failed or had poor empirical performance. Iteratively attending to the inputs is more computationally efficient, scalable and more consistent statistically than the other approaches. 

\subsection{Attention Mechanism}
The attention mechanism we used is incrementally derived from zoom-able and differentiable image observation mechanism of DRAW ~\citet{gregor2015draw}. The attention window is defined by an $N \times N$ 2D grid of Cauchy kernels. We found that the heavy tail of the Cauchy curve alleviates some of the vanishing gradient issues and it also increases the speed of training.

The grid's location and size is defined based on the glimpse parameters. The $N \times N$ grid of kernels is placed at $(x, y)$ on the $S \times S$ image, with the central Cauchy kernel being located at $(x, y)$. The elemental square in the grid has a side of length $\delta$. The glimpse parameter set $\Omega_t$ is unpacked to get $\Omega_t \rightarrow (\widehat{x},\widehat{y},\widehat{\delta})$. $x, y$ and $\delta$ are computed from  $\widehat{x},\widehat{y}$ and,$\widehat{\delta}$ using the following transforms:

\begin{center}
$x = (S-1) \: \frac{(\widehat{x} + 1)}{2}\quad$ 
$y = (S-1) \: \frac{(\widehat{y} + 1)}{2}\;$
\vskip 0.1in
$\delta = \frac{S}{N} (1 - \vert\widehat{\delta}\vert)\quad$
$\gamma = e^{1 - 2\vert\widehat{\delta}\vert}$
\end{center}

The location of a $i^{th}$ row, $j^{th}$ column's Cauchy kernel in terms of the pixel coordinates of the image is given by:

$\mu_X^i = x + (i - (N + 1)/2)\:\delta \quad$
$\mu_Y^j = y + (j - (N + 1)/2)\:\delta$

The horizontal and vertical filterbank matrices are then calculated as:
\begin{center}
$F_{X}[i, a] = \frac{1}{Z_X}\left\lbrace {\pi \gamma \left[ 1 + \left( \frac{a - \mu_X^i}{\gamma}\right) ^{2} \right]}\right\rbrace ^{-1} \; $
$
F_{Y}[j, b] = \frac{1}{Z_Y}\left\lbrace {\pi \gamma \left[ 1 + \left( \frac{b - \mu_Y^j}{\gamma}\right) ^{2} \right]}\right\rbrace ^{-1}$
\end{center}

${Z_X}$ and ${Z_Y}$ are normalization constants such that they make $\Sigma_a F_{X}[i, a] = 1$ and $\Sigma_b F_{X}[j, b] = 1$

Final result of attention on an image is given by:
\begin{center}
$attend(I_t, \Omega_t) = F_YI_tF_X^T$
\end{center}

$attend$ thus gets an $N \times N$ patch of the image, which is flattened and used in the model.

\subsection{Use of Convolutions}
As seen in the experimental sections that follow, use of convolutional feature extractors gave a significant boost in performance. Given an image, the application of several layers of convolution produces a 3D solid of activations (or a stack of 2D feature maps). Attention over this corresponds to applying the same 2D attention (described in Section 2.1 above) over the entire depth of the 3D feature map. The attended sub-solid is then flattened and used as the glimpse.

\section{Dynamic Representations and One-shot Classification}
\label{omniglot-one-shot}
One-shot learning requires learning models to be at the apotheosis of data efficiency. In the case of one-shot classification, only a single example of each individual class is given and the model is expected to generalise to new samples of the same class.

\subsection{Dynamic Representations}
Deep Neural Networks learn useful representations of objects from data. Representation of a sample is computed by identifying a fixed set of features in it, and these features are learnt from scratch and are purely based on data provided during training. In the end, a trained neural network can be interpreted as being composed of two components - a function that maps the input sample to a point in representation space and a classifier that learns a decision boundary in this representation space. 

Rapid learning expects that this representation space to be dynamic -- representations should change with newly found data. All features that form a good representation aren't known during initial learning and entirely new concepts with never-before-seen features should be expected. Ideally, the entire representation space should change when the new concept is introduced. This would enable the assimilation of new concepts in conjunction with the old concepts. One way of training such systems is to have a meta-learning system where the model is trained to represent entities in space (rather than being trained to represent an entity) ~\citep{schaul2010metalearning}. Deep Learning research in this direction recently  ~\citep{santoro2016one} has explored developing complex models that are trained in an end-to-end manner. But empirically, we found that such top-down hierarchical models are difficult to train, suffer from reduced supervision and require specially constructed datasets.

However, there is another alternative strategy that could be employed as crude approximation of this ideal scenario. This involves lazily developing a representation space that is conditioned on the test sample only at inference time. Until then, all samples presented to the model are just stored as-is in a repository. When the test sample is given, we compare this sample with every other sample in our repository using ARCs to form a relative representation of each sample (the representation being the final hidden state of the recurrent controller). In this relative representation space, which is relative to a test sample, we use a trained classifier that can identify the most similar sample pair, given the entire context of relative representation space. This relative representation space is dynamic as it changes relative to the test sample. 

\subsection{One-shot Learning Models}
The standard one-shot classification setup consists of a support set and a test sample. In an one-shot learning episode, the support set containing a single example of each class is first provided to the model. Next, a test sample is given and the model is expected to make its classification prediction. Finally, the classification accuracy is calculated based on all the predictions. We developed the following two models with ARCs for this task:

\subsubsection{Naive ARC Model}
This is a trivial extension of ARCs for used for the verification task. A test sample is compared against all the images in the support set. It is matched to the sample with maximum similarity and the corresponding class is the prediction of the model. Here, we are reducing the relative representations to similarity scores directly. The entire context of the relative representation space is not incorporated.

\subsubsection{Full Context ARC}
This model incorporates the full knowledge of the relative representation space generated before making a prediction. While Naive ARC model is simple and efficient, it does not incorporate the whole context in which our model is expected to make the decision of similarity. When the test sample is being compared with all support samples, the comparisons are all independently done. 

It is highly desirable to have a 20-way ARC, where each observation is conditioned on the all images in the background set. Unfortunately, such a model is not practical. This would require maintaining a lot of context in the controller state. But, scaling up the controller memory incurs a huge (quadratic) parameter burden. So instead, we use a hierarchical setup, which decomposes the comparisons to be at two levels - first local pairwise comparison and a second global comparison. We found that this model reduces the information that has to be crammed in the controller state, while still providing sufficient context.

As with the Naive method, we compare test sample from evaluation set with each image from support set in pairs. But instead of emitting a similarity score immediately, we process the \textit{relative representations} of each comparison. Relative representations are the final hidden state of the controller when the test image $T$ is being compared to image $S_j$ from the support set: $e_j = {h_L}^{T, S_j}$. These embeddings are further processed by a Bi-Directional LSTM layer. This merges the information from all comparisons, thus providing the necessary context before prediction. The approach taken here is very similar to Matching Networks ~\citep{vinyals2016matching}, but it is slightly more intuitive and provides superior results.

\begin{center}
$c_j = [\:\overrightarrow{LSTM}(e_j); \; \overleftarrow{LSTM}(e_j)\:] \qquad \forall j \in [1, 20]$
\end{center}

Each embedding is mapped to a single score $s_j = f(c_j)$, where $f(.)$ is an affine transform followed by a non-linearity. The final output is the normalized similarity with respect to all similarity scores. 
\begin{center}
$p_j = softmax(s_j) \qquad \forall j \in [1, 20]$
\end{center}

This whole process is to make sure that we adhere to the fundamental principle of Deep Learning, which is to optimise objectives that directly reflect the task. The softmax normalisation allows for the expression of relative similarity rather than absolute similarity. 

\section{Experiments}
In this section, we first detail the analysis done to better understand the empirical functioning of ARCs, both qualitatively and quantitatively. We then benchmark ARCs on standard similarity learning tasks on two datasets and present the results.

\subsection{Model Analysis}
For the analysis presented below, we use the simple ARC model described in Section 2 trained for the verification (or similarity learning) task on the Omniglot dataset. The verification task is a binary classification problem wherein the model is trained to predict whether the 2 drawings provided are of the same character or not. 

The final hidden state of the RNN controller $h_T$ is used to output at a single logistic neuron that estimates the probabilty of similarity. The particular model under consideration has an LSTM controller ~\citep{hochreiter1997long} with forget gates ~\citep{gers2000learning}. The number of glimpses per image was fixed to 8, thus making the total number of recurrent steps 16. $32\times32$ greyscale images of characters were used and the attention glimpse resolution of $4\times4$ was used. Similar/dissimilar pairs of character drawings were randomly chosen from within the same language to make the task more challenging. 

\subsubsection{Omniglot Dataset}
Omniglot is a dataset by ~\cite{lake2015human} that is specially designed to compare and contrast the learning abilities of humans and machines. The dataset contains handwritten characters of 50 languages (alphabets) with 1623 total characters. The dataset is divided into a background set and an evaluation set. Background set contains 30 alphabets (964 characters) and only this set should be used to perform all learning (e.g. hyper-parameter inference or feature learning). The remaining 20 alphabets are for pure evaluation purposes only. Each character is a $105 \times 105$ greyscale image. There are only 20 samples for each character, each drawn by a distinct individual.

\begin{figure}
\centering     
\subfigure[It can be seen that the two characters look very similar in their stroke pattern and differ only in their looping structure. ARC has learnt to focus on these crucial aspects.]{\label{fig:a}\includegraphics[width=80mm]{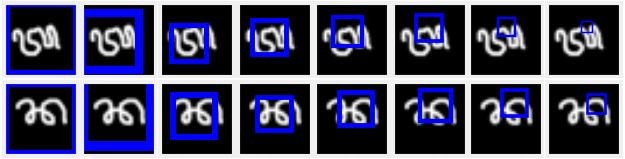}}
\subfigure[ARC parses over the characters in a left to right, top to bottom fashion. Finally, it ends up focussing in the region where the first character has a prolonged downward stroke, whereas the second one does not.]{\label{fig:b}\includegraphics[width=80mm]{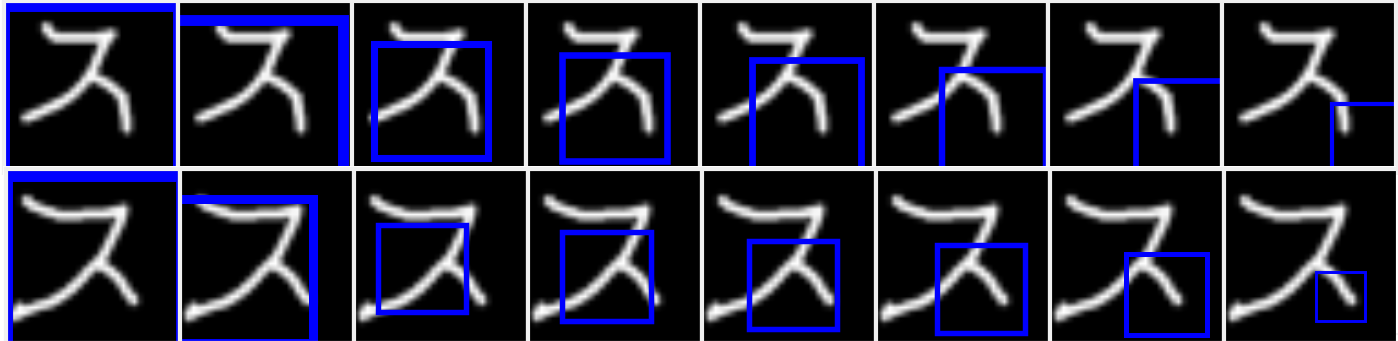}}
\caption{Attention windows over time when comparing the two Omniglot characters. The top row has the first image and the bottom row has the second. Each column represents a glimpse step. (a) Comparing two dissimilar characters and (b) Comparing two similar characters.}
\end{figure}

\subsubsection{Qualitative Analysis}
ARCs tend to adopt a \textit{left to right} parsing strategy for most pairs, during which the attention window gradually reduces in size. As seen in Figures 2(a) and 2(b), the observations made by ARC in one image are definitely being conditioned on the observations in the other image. We also frequently encountered cases wherein the attention window, would end up focusing on a blank region.

\subsubsection{Quantitative Analysis}
We performed simple yet insightful ablation studies to understand ARC's dynamics. The ARC accumulates information about both the input images by a series of attentive observations. To see how the information content varied with observations, we trained 8 separate binary classifiers to classify images as being similar or not based on hidden states of the LSTM controller at every even time-step. The performance of these classifiers is summarized in Table 1. Since the ARC has an attention window of only $4\times4$ pixels, it can barely see anything in the first time step, where its attention is spread throughout the whole image. With more glimpses, finer observations bring in more precise information and the recurrent transitions make use of this knowledge, leading to higher accuracies. We also used the 8 binary classifiers to study how models confidence grows with more glimpses and such examples are provided in Figure 3.

\begin{table}[h]
\caption{Glimpses per image versus classification accuracy of ARC. Out of the 50 alphabets provided in the Omniglot dataset, 30 were used for training and validation and the last 20 for testing}
\label{sample-table}
\vskip 0.15in
\begin{center}
\begin{small}
\begin{sc}
\begin{tabular}{cc}
\hline
\abovespace\belowspace
Glimpses & Accuracy (Test Set) \\
\hline
\abovespace
1 &58.2\%\\
2 &65.0\%\\
4 &80.8\%\\
6 &89.25\%\\
\textbf{8} &\textbf{92.08\%}\\
\hline
\end{tabular}
\end{sc}
\end{small}
\end{center}
\vskip -0.1in
\end{table}

\begin{figure}
\centering     
\subfigure[ARC is very unsure of similarity at the beginning. But at 5th glimpse (4th column), the attention goes over the region where there are strokes in the first image and no strokes in the second one resulting in dropping of the score.]{\label{fig:a}\includegraphics[width=80mm]{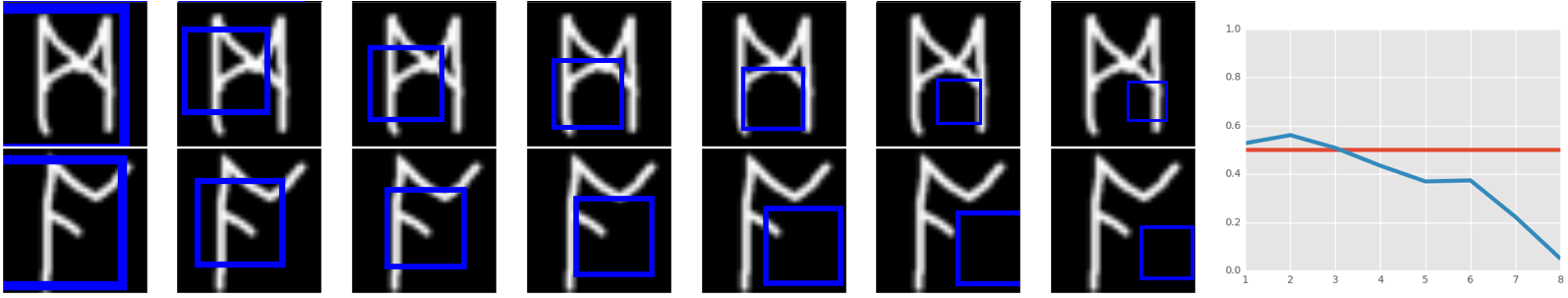}}
\subfigure[Initially ARC is unsure or thinks that the characters are similar. But towards the end, at 6th glimpse (5th column), the model focusses on the region where the connecting strokes are different. The similarity score drops and with more "ponder", it falls down significantly.]{\label{fig:b}\includegraphics[width=80mm]{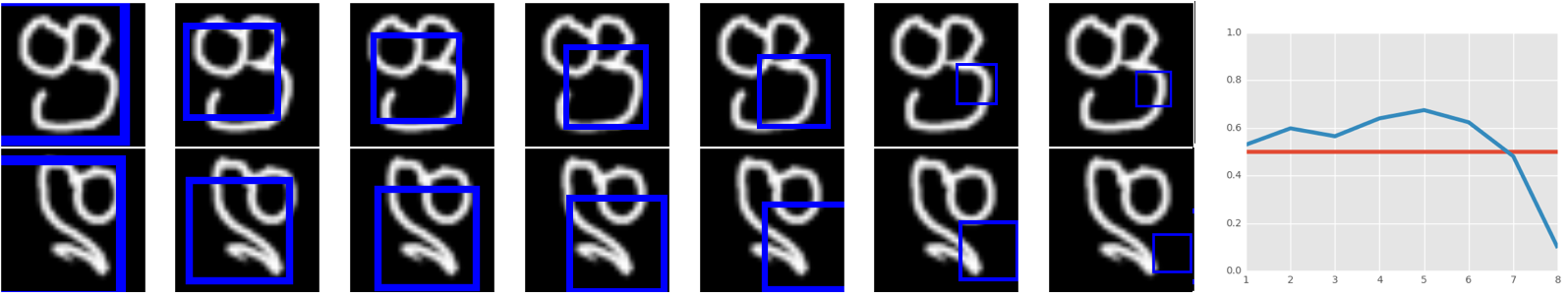}}
\caption{Attention windows over time and instantaneous predictions from independent binary classifiers. The first glimpse is omitted as it covers the whole image. In the graph: x-axis: glimpse number, y-axis: similarity score. The red line is the decision threshold, above which the images are considered to be similar. Both of the cases above are examples of a dissimilar pair.}
\end{figure}

\subsection{Similarity Learning}
In this section we compare ARCs with other Deep Learning methods in the task of similarity learning on two datasets: Omniglot and CASIA WebFace Dataset. We consider strong convolutional baselines, which have been shown time and again to excel at such visual tasks. Particularly, we use Wide Resnets (WRNs) ~\citep{DBLP:journals/corr/ZagoruykoK16} which are the current state of the art models in image classification. Wide ResNets used contain 4 blocks of convolutional feature extractors. ConvARC models also used Wide Resnets for feature extraction but with one less block of convolutions. We used moderate data augmentation consisting of translation, flipping, rotation and shearing, which we found to be critical for training ARC models (WRNs also were trained with the same augmentation).  Hyper parameters were set for reasonable values for all our ARC models and no hyper-parameter tuning of any kind was employed. 

\subsubsection{Omniglot}
The same exact model used in the previous section was used for this comparison as well. The data split up of the Omniglot dataset used for this comparison is different from the above: 30 alphabets were used for training, 10 for validation and 10 for testing (this was in order to be comparable to the ConvNets in ~\citep{kochsiamese}).The results are aggregated in Table 2. Our simple ARC model without using any convolutional layers obtains a performance that matches a AlexNet style 6 layer Deep Convnet. Using convolutional feature extractors, ARCs outperform the Wide ResNet based Siamese ConvNet baselines, even the ones containing an order of magnitude more parameters.

\subsubsection{CASIA Webface}
CASIA Webface is the largest public repository of faces consisting of 494,414 distinct images of over 10 thousand people. We split the data as follows: Training set: 70\% (7402 people), validation set: 15\% (1586 people) and Test set: 15\% (1587 people). The images were downscaled to 32x32 pixels and an attention window of 4x4 pixels was used. The rest of the architecture is same as the Omniglot model. Results are tabluated in Table 3.

\begin{table}[t]
\caption{Performance of ARC vs conventional methods on the verification task on Omniglot dataset. Wide ResNets suffixes specify the depth and width. Example, \textit{(d=60, w=4)}  means that it is a ResNet that 60 is layers deep with each residual block having a width multiplier of 4. Out of the 50 alphabets provided in the Omniglot dataset, 30 were used for training, 10 for validation and the last 10 for testing}
\label{sample-table}
\vskip 0.15in
\begin{center}
\begin{small}
\begin{sc}
\begin{tabular}{cc}
\hline
\abovespace\belowspace
Model & Accuracy (Test Set)\\
\hline
\abovespace
Siamese Network        &60.52\% \\
Deep Siamese Net~\citep{kochsiamese}   &93.42\%\\
Siamese ResNet (d=24, w=1)   &93.47\%\\
Siamese ResNet (d=30, w=2)   &94.61\%\\
Siamese ResNet (d=60, w=4)   &93.57\%\\
\textbf{ARC}    &\textbf{93.31\%}\\
\textbf{ConvARC}    &\textbf{96.10\%}\\
\hline
\end{tabular}
\end{sc}
\end{small}
\end{center}
\vskip -0.1in
\end{table}

\begin{table}[t]
\caption{Performance of ARC vs conventional methods on the verification task on CASIA Webface dataset. Wide ResNets suffixes notation is same as used in Table 2.}
\label{sample-table}
\vskip 0.15in
\begin{center}
\begin{small}
\begin{sc}
\begin{tabular}{cc}
\hline
\abovespace\belowspace
Model & Accuracy (Test Set)\\
\hline
\abovespace
Siamese ResNet (d=36, w=4)   &79.48\%\\
ARC   &72\%\\
\textbf{ConvARC}    &\textbf{81.73\%}\\
\hline
\end{tabular}
\end{sc}
\end{small}
\end{center}
\vskip -0.1in
\end{table}

\section{One Shot Classification}
One-shot classification on the Omniglot dataset is a very challenging task as most Deep Learning systems do not work well on this dataset. ~\cite{lake2015human} developed a dedicated system for such rapid knowledge acquisition called Bayesian Programming Learning, which surpasses human performance and is the current state of the art method.

The details of the Omniglot dataset are given in Section 4.1.1 . One-shot classification task on this dataset is setup as follows: from a randomly chosen alphabet, 20 characters are chosen which becomes the support set classes. One character among these 20 becomes the test character. 2 drawers are chosen, one each for the support set drawings and the test character drawing. The task is to match the test drawing to the correct character's class in the support set. Assigning an image to one of the 20 characters results in a 20-way, 1-shot classification task.

\subsection{Baselines and Other Methods}
We compare the two models discussed in Section 3.2 with other methods in literature: starting from the simplest baseline of k-Nearest Neighbours to the latest meta-learning methods. The training and evaluation practices are non-consistent and the two main threads of variation are detailed below. 

\textbf{Across Alphabets}: Many papers recently, like Matching Networks ~\cite{vinyals2016matching} and MANNs ~\cite{santoro2016one} have used 1200 chars for the background set (instead of 964 specified by ~\cite{lake2015human}). The remaining 423 characters are used for testing. Most importantly, the characters sampled for both training and evaluation are across all the alphabets in the training set.

\textbf{Within Alphabets}: This corresponds to standard Omniglot setting where characters are sampled within an alphabet and only the 30 background characters are used for training and validation.

The across alphabet task is much simpler as it is easy to distinguish characters belonging to different languages, compared to distinguishing characters belonging to the same language. 

There are large variations in the resolution of the images used as well. The Deep Siamese Network of ~\citet{kochsiamese} uses 105x105 images and thus not directly comparable to out model, but we include it as it is the current best result using deep neural nets. The performance of MANNs in this standard setup is interpreted from the graph in the paper as the authors did not report it. It should also be noted that Bayesian Program Learning (BPL) ~\cite{lake2015human} incorporates human stroke data into the model. Lake et al estimate the human performance to be at 95.5\%. 

Results are presented in Table 4 and 5. Our ARC models outperform all previous methods according to both of the testing protocols and establish the corresponding state of the art results.

\begin{table}[!h]
\caption{One-shot classification accuracies of various methods and our ARC models on Omniglot dataset - Across Alphabets}
\label{sample-table}
\vskip 0.15in
\begin{center}
\begin{small}
\begin{sc}
\begin{tabular}{lc}
\hline
\abovespace\belowspace
MODEL  & ACCURACY
\\ \hline \\
			kNN 	&26.7\%\\
			Conv Siamese Network         &88.1\% \\	
			MANN         &$\approx$ 60\%\\
			Matching Networks       &93.8\%\\
			Naive ARC     &90.30\% \\
			\textbf{Naive ConvARC}    &\textbf{96.21\%} \\
			\textbf{Full Context ConvARC}     &\textbf{97.5\%} \\
\hline
\end{tabular}
\end{sc}
\end{small}
\end{center}
\vskip -0.1in
\end{table}

\begin{table}[!h]
\caption{One-shot classification accuracies of various methods and our ARC models on Omniglot dataset - Within Alphabets}
\label{sample-table}
\vskip 0.15in
\begin{center}
\begin{small}
\begin{sc}
\begin{tabular}{lc}
\hline
\abovespace\belowspace
MODEL  & ACCURACY
\\ \hline \\
			kNN	&21.7\%\\
			Siamese Network         &58.3\% \\					
			Deep Siamese Network~\citep{kochsiamese}    &92.0\%\\
			Humans &95.5\%\\
			BPL &96.7\%\\
			Naive ARC     &91.75\% \\
			\textbf{Naive ConvARC}    &\textbf{97.75\%} \\
			\textbf{Full Context ConvARC}     &\textbf{98.5\%} \\ \\
\hline
\end{tabular}
\end{sc}
\end{small}
\end{center}
\vskip -0.1in
\end{table}

\subsection{miniImageNet}
Recently, \citet{vinyals2016matching} introduced a one-shot learning benchmark based off of the popular ImageNet dataset. It uses a testing protocol that is very similar to Omniglot. The dataset consists of  60,000 colour images of size $84 \times 84$ with 100 classes of 600 examples each. As with the original paper, we used 80 classes for training and tested on the remaining 20 classes. We report results on 5-way one-shot task in Table 6, which is a one-shot learning with 5 classes at a time.

\begin{table}[!h]
\caption{5 way one-shot Classification accuracies of various methods and our ARC models - miniImageNet}
\label{sample-table}
\vskip 0.15in
\begin{center}
\begin{small}
\begin{sc}
\begin{tabular}{lc}
\hline
\abovespace\belowspace
MODEL  & ACCURACY
\\ \hline \\
Raw Pixels w/ Cosine Similarity	&23.0\%\\
Baseline Classifier        &38.4\% \\					
Matching Networks &46.6\%\\  
\textbf{Naive ConvARC}    &\textbf{49.14\%}\\
\hline
\end{tabular}
\end{sc}
\end{small}
\end{center}
\vskip -0.1in
\end{table}

\section{Related Work}
\label{related}

Deep Neural Networks~\citep{schmidhuber2015deep}~\citep{lecun2015deep} are very complex parametrised functions which can be adapted to have the required behaviour by specifying a suitable objective function. Our overall model is a simple combination of the attention mechanism and recurrent neural networks (RNNs).

It is known that attention brings in selectivity in processing information while reducing the processing load~\citep{desimone1995neural}. Attention and (Recurrent) Neural Networks were combined in~\citet{schmidhuber1991learning} to learn fovea trajectories. Later attention was used in conjunction with RBMs to learn what and where to attend in~\citet{larochelle2010learning} and in~\citet{denil2012learning}. Hard Attention mechanism based on Reinforcement Learning was used in~\citet{mnih2014recurrent} and further extended to multiple objects in~\citet{ba2014multiple}; both of these models showed that the computation required at inference is significantly less compared to highly parallel Convolutional Networks, while still achieving good performance. A soft or differentiable attention mechanisms have been used in ~\citet{graves2013generating}. A specialised form of location based soft attention mechanism, well suited for 2D images was developed for the DRAW architecture~\citep{gregor2015draw}, and this forms the basis of our attention mechanism in ARC.

A survey of the methods and importance of measuring similarity of samples in Machine Learning is presented in~\citet{bellet2013survey}. With respect to Deep Learning methods, the most popular architecture family is that of Siamese Networks~\citep{bromley1993signature}. The energy based derivation of the same is presented in~\citet{chopra2005learning}. 

A bayesian framework for one-shot visual recognition was presented in~\citet{fe2003bayesian}. ~\citet{lake2015human} extensively study one-shot Learning and present a novel probabilistic framework called Bayesian Program Learning (BPL) for rapid learning. They have also released the Omniglot dataset, which has become a testing ground for one-shot learning techniques. Recently, many Deep Learning methods have been developed to do one-shot learning:~\citet{kochsiamese} use Deep Convolutional Siamese Networks for performing one-shot classification. Matching Networks~\citep{vinyals2016matching} and Memory Augmented Neural Networks~\citep{santoro2016one} are other approaches to perform continual or meta learning in the low data regime.

\section{Conclusion and Future Work}
We presented a model that uses attention and recurrence to cycle through a set of images repeatedly and estimate their similarity. We showed that this model is not only viable but is also much better than the popular siamese neural networks in wide use today in terms of performance and generalization. We showed the value of having dynamic representations and presented a novel way of approximating it. Our main result is in the task of one-shot classification on the Omniglot dataset, where we achieved state of the art performance surpassing human performance using only raw pixel data.

Though presented in the context of images, ARCs can be used for  any modality. There are innumerable ways to extend ARCs. Better attention mechanisms, higher resolution images, careful hyper-parameter tuning, more complicated controllers etc ., can be employed to achieve better performance. However, one potential downside of this model is that due to sequential execution of the recurrent core and by the very design of the model, it might be more computationally expensive than distance metric methods.

More interesting directions would involve developing more complex architectures using this bottom-up, lazy approach to solve even more challenging AI tasks.
 
\section*{Acknowledgements} 
We would like to thank Akshay Mehrotra, Gaurav Pandey and Siddharth Agrawal for their extensive support and feedback while developing the ideas in this work. We would like to thank Soumith Chintala for his feedback on this manuscript and the
idea and Ankur Handa for helping us with the CASIA dataset. We would also like to thank Sanyam Agarwal, Wolfgang Richter, Shreyas Karanth and Aadesh Bagmar for their valuable feedback on the manuscript. Finally, we would like to thank the anonymous reviewers for helping us significantly improve the quality of this paper.

Authors acknowledge financial support for the "CyberGut" expedition project by the Robert Bosch Centre for Cyber Physical Systems at the Indian Institute of Science, Bengaluru.

\bibliography{example_paper}
\bibliographystyle{icml2017}

\end{document}